\documentclass{article} % For LaTeX2e
\usepackage{amssymb}

\usepackage[preprint]{corl_2025}

\usepackage{microtype}
\usepackage{hyperref}
\usepackage{url}
\usepackage{booktabs}
\usepackage{graphicx}
\usepackage{lineno}
\usepackage{titletoc}
\usepackage[dvipsnames]{xcolor}

\definecolor{darkblue}{rgb}{0, 0, 0.5}
\hypersetup{colorlinks=true, citecolor=darkblue, linkcolor=darkblue, urlcolor=darkblue}

\usepackage{epsfig}
\usepackage{graphicx}
\usepackage{float}
\usepackage{wrapfig}
\usepackage{algorithm,algorithmicx,algpseudocode}
\usepackage{bm,xspace}
\usepackage{comment}
\usepackage{verbatim}
\usepackage{multirow}
\usepackage{array}
\usepackage{balance}
\usepackage{url}
\usepackage{booktabs}
\usepackage{etoolbox,siunitx}
\usepackage{calc}
\usepackage{pifont,hologo}
 
\usepackage{nicefrac}

\usepackage{arydshln}

\usepackage[utf8]{inputenc}
\UseRawInputEncoding
\usepackage[T1]{fontenc}
\usepackage{url} 
\usepackage{amsfonts}
\usepackage{nicefrac}
\usepackage{microtype}
\usepackage[american]{babel}
\usepackage{enumitem}
\usepackage{amssymb}
\usepackage{amsmath}
\usepackage{siunitx}
\usepackage{bbm}

\usepackage[super]{nth}
\usepackage{nicefrac}
\robustify\bfseries

\usepackage{dsfont}

\newcommand{\sA}{\mathcal{A}}

\newcommand{\sD}{\mathcal{D}}

\newcommand{\sS}{\mathcal{S}}

\DeclareMathSymbol{@}{\mathord}{letters}{"3B}

\makeatletter
\DeclareRobustCommand\onedot{\futurelet\@let@token\@onedot}
\def\@onedot{\ifx\@let@token.\else.\null\fi\xspace}

 \def\Eg{\emph{A.k.a}\onedot}
\def\eg{\emph{e.g}\onedot} \def\Eg{\emph{E.g}\onedot}
\def\ie{\emph{i.e}\onedot} 
\def\cf{\emph{c.f}\onedot} 
 \def\vs{\emph{vs}\onedot}

\makeatother

\usepackage{algorithm,algorithmicx,algpseudocode}
\usepackage{pifont}% http://ctan.org/pkg/pifont
\usepackage{xspace}
\newcommand{\cmark}{\ding{51}}%
\newcommand{\xmark}{\ding{55}}%

\newcommand{\alfred}{{ALFRED}\xspace}
\newcommand{\rtfm}{{RTFM}\xspace}
\newcommand{\overcook}{{Overcooked-AI}\xspace}

\newcommand{\planner}{\textsc{Fast-Downward}\xspace}

\newcommand{\ourmethod}{\textrm{Predicting Semantics of Actions with Language Models for Vision and Robotics}\xspace}

\newcommand{\ourmethodshort}{\textrm{{PSALM-V}}\xspace}

\title{PSALM-V: Automating Symbolic Planning in Interactive Visual Environments with \\ Large Language Models}

\author{%
  Wang Bill Zhu \quad\quad Miaosen Chai \quad\quad Ishika Singh \vspace{4pt} \\ \textbf{Robin Jia \quad\quad Jesse Thomason} \vspace{4pt} \\
  Department of Computer Science \\
  University of Southern California\\
  $^\dagger$Corresponding author: \texttt{wangzhu@usc.edu} \\
}

\begin{document}

\maketitle

\begin{abstract}
We propose \ourmethodshort, the first autonomous neuro-symbolic learning system able to induce symbolic \emph{action semantics (i.e., pre- and post-conditions)} in visual environments through interaction.
\ourmethodshort\ bootstraps reliable symbolic planning without expert action definitions, using LLMs to generate heuristic plans and candidate symbolic semantics.
Previous work has explored using large language models to generate action semantics for Planning Domain Definition Language (PDDL)-based symbolic planners. 
However, these approaches have primarily focused on text-based domains or relied on unrealistic assumptions, such as access to a predefined problem file, full observability, or explicit error messages.
By contrast, \ourmethodshort dynamically infers PDDL problem files and domain action semantics by analyzing execution outcomes and synthesizing possible error explanations. 
The system iteratively generates and executes plans while maintaining a tree-structured belief over possible action semantics for each action, iteratively refining these beliefs until a goal state is reached.
Simulated experiments of task completion in \alfred demonstrate that \ourmethodshort increases the plan success rate from 37\% (Claude-3.7) to 74\% in partially observed setups. 
Results on two 2D game environments, \rtfm and \overcook, show that \ourmethodshort improves step efficiency and succeeds in domain induction in multi-agent settings.
\ourmethodshort correctly induces PDDL pre- and post-conditions for real-world robot BlocksWorld tasks, \emph{despite low-level manipulation failures from the robot}.
Videos and resources at \url{https://psalmv.github.io/}.
\end{abstract}


\section{Introduction}
\label{sec:intro}

Planning in interactive, visual environments presents a fundamental challenge for embodied artificial intelligence. 
Symbolic planners have traditionally been a powerful tool for structured decision-making, generating sequences of actions to transition from an initial state to a goal state. 
These planners rely on expert-defined action semantics, typically encoded in domain-specific languages such as the Planning Domain Definition Language (PDDL;~\cite{aeronautiques1998pddl}). 
Constructing these logical action semantics requires extensive manual effort and domain knowledge, limiting scalability to new environments.

Recent LLM and robot research has explored leveraging large language models (LLMs) to alleviate the need for human-authored action semantics by generating PDDL-like action definitions~\citep{zhu2024languagemodelsinferaction, Wong2023LearningAP}. 
While promising, these existing approaches have strong and unrealistic assumptions, such as:
\begin{itemize}[label={$\bullet$}, left=0pt, topsep=0pt, itemsep=0pt]
    \item \textbf{Focus on text-based domains}: Real-world scenarios often require visual interaction to understand environmental dynamics.
    \item \textbf{Reliance on full observability}: Many real-world scenarios involve partial observability, where agents must infer missing information from past interactions.
    \item \textbf{Predefined, fully observed symbolic initial state}: While PDDL problem files characterizing the initial state of the world are provided in some environments, generating them, particularly in visual environments, is complex.
    \item \textbf{Explicit error messages}: Error messages are usually not available in real-world exploration. Previous work by \citet{zhu2024languagemodelsinferaction} highlights failures without explicit error messages.
    \item \textbf{List structure of pre- and post-conditions}: The assumption that action semantics are intersections of logical conditions joined by \texttt{and} limits the use of other logical operators like \texttt{or} and \texttt{when}.
\end{itemize}

To address these limitations, we introduce \ourmethod (\ourmethodshort), the first neuro-symbolic framework that tackles key challenges in automated planning through flexible semantic updates and dynamic trajectory sampling. Starting with a problem file initialized from visual input, \ourmethodshort explores partially observed environments, constructing semantic maps to aid problem file generation. The method iteratively samples and executes trajectories, predicts error messages, and updates action semantics to refine memory. It maintains a tree-structured belief over possible action semantics for each action, refining these beliefs until a valid plan is executed to achieve the goal state. 
A PDDL domain file, a symbolic representation of the rules of the world, is sampled based on the current memory of hypothesized action semantics, and a symbolic planner searches for a path to the goal state given those rules.

We evaluate \ourmethodshort in three visual simulation environments and in a physical manipulation task with a Franka Panda arm.
In simulation, we explore \alfred~\citep{shridhar2020alfred}, \rtfm~\citep{zhong2021rtfmgeneralisingnovelenvironment}, and \overcook~\citep{carroll2019utility}. 
In \alfred, \ourmethodshort increases the planning success rate from 37\% (Claude-3.7) to 74\% under conditions of partial observability. In \rtfm and \overcook, \ourmethodshort\ improves step efficiency and successfully induces action semantics in multi-agent settings. 
Real-world BlocksWorld robot experiment results show that \ourmethodshort successfully recovers the full domain (F1 of 100\%) with an average of 9.3 environment steps and achieves a goal-conditioned completion rate of 66.7\%.

\begin{figure}
    \centering
    \includegraphics[width=\linewidth]{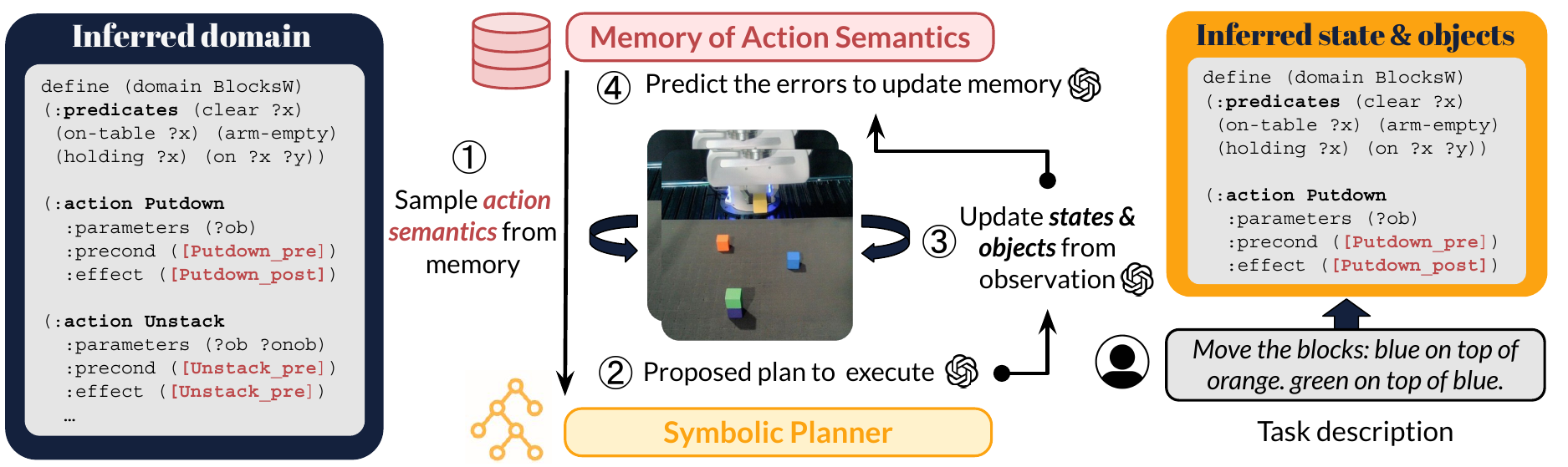}
    \caption{LLMs can propose plans and generate action semantics, but struggle with state tracking. PDDL symbolic planners leverage specialized search algorithms, but require predefined action semantics for the environment. \ourmethodshort integrates the strengths of both for tasks on partially observed, visual environments.}
    \label{fig:teaser}
\end{figure}
\begin{table*}[ht]
\centering
% \small
\resizebox{\linewidth}{!}{
\begin{tabular}{lcccccccc}
    & \bf Objective & \multicolumn{5}{c}{\bf Independent of} & {\bf Visual \& Robot} \\
    \cmidrule(lr){3-7}
    & & Partial AS & Valid plans & Human eval & PF & Full state & \\
    \toprule
    \citet{llmp} & PF & {\color{BrickRed} \xmark} & {\color{blue} \cmark} & {\color{blue} \cmark} & {\color{BrickRed} \xmark} & {\color{BrickRed} \xmark} & {\color{BrickRed} \xmark} \\
    \citet{zhang-etal-2024-pddlego} & PF & {\color{blue} \cmark} & {\color{blue} \cmark} & {\color{blue} \cmark} & {\color{BrickRed} \xmark} & {\color{blue} \cmark} & {\color{BrickRed} \xmark} \\
    \citet{lin2024clmasp} & Plan & {\color{blue} \cmark} & {\color{blue} \cmark} & {\color{BrickRed} \xmark} & - & {\color{BrickRed} \xmark} & {\color{BrickRed} \xmark} \\
    \citet{Silver2023GeneralizedPI} & Program & {\color{BrickRed} \xmark} & {\color{blue} \cmark} & {\color{blue} \cmark} & {\color{blue} \cmark} & {\color{BrickRed} \xmark} & {\color{BrickRed} \xmark} \\
    \midrule
    \citet{arora:hal-02010536} & AS & {\color{blue} \cmark} & {\color{BrickRed} \xmark} & {\color{blue} \cmark} & - & {\color{BrickRed} \xmark} & {\color{BrickRed} \xmark} \\ 
    \citet{Wong2023LearningAP} & AS & {\color{BrickRed} \xmark} & {\color{blue} \cmark} & {\color{blue} \cmark} & {\color{blue} \cmark} & {\color{BrickRed} \xmark} & {\color{BrickRed} \xmark} \\ 
    \citet{guan2023leveraging} & AS & {\color{blue} \cmark} & {\color{blue} \cmark} & {\color{BrickRed} \xmark} & {\color{blue} \cmark} & {\color{BrickRed} \xmark} & {\color{BrickRed} \xmark}  \\
    \citet{oswald2024large} & AS & {\color{blue} \cmark} & {\color{blue} \cmark} & {\color{blue} \cmark} & {\color{BrickRed} \xmark} & {\color{BrickRed} \xmark} & {\color{BrickRed} \xmark} \\
    \citet{zhu2024languagemodelsinferaction} & AS & {\color{blue} \cmark} & {\color{blue} \cmark} & {\color{blue} \cmark} & {\color{BrickRed} \xmark} & {\color{BrickRed} \xmark} & {\color{BrickRed} \xmark} \\
    \citet{han2024interpret} & {AS} & {{\color{blue} \cmark}} & {{\color{blue} \cmark}} & {{\color{BrickRed} \xmark}} & {\color{blue} \cmark} & {\color{BrickRed} \xmark} & {\color{blue} \cmark} \\
    \midrule
    \ourmethodshort & AS & {\color{blue} \cmark} & {\color{blue} \cmark} & {\color{blue} \cmark} & {\color{blue} \cmark} & {\color{blue} \cmark} & {\color{blue} \cmark} \\
    \bottomrule        
\end{tabular}
}
\caption{\ourmethodshort domain induction task setup \vs representative related works in LLM and LLM-Modulo planning; here, ``AS'' is short for action semantics, ``PF'' is short for problem file, ``Full state'' means full state info is given.}
\label{tab:rw_table}
\end{table*}

\section{Background and Related Works}
\label{sec:related}

We first introduce symbolic planning, and then compare \ourmethodshort's domain induction setup with previous LLM-augmented planning approaches, enumerated in Table~\ref{tab:rw_table}.

\subsection{Classical and Symbolic Planning}
\label{subsec:classical}
Classical planning algorithms have been extensively applied in domains such as autonomous spacecraft control, military logistics, manufacturing, games, and robotics. 
The automated STRIPS planner was the first algorithm to operate the Shakey robot~\cite{strips}. 
Classical planners assume a finite, deterministic environment with full state observability, enabling them to generate guaranteed plans when a path from the initial to the goal state exists. Other planning frameworks, such as PRODIGY~\cite{prodigy} and HTNs~\cite{htn}, have also proven effective in robotic planning contexts.
Symbolic planning languages, such as the Planning Domain Description Language (PDDL;~\cite{aeronautiques1998pddl}) and Answer Set Programming (ASP;~\cite{asp, asp1}), offer structured and expressive ways to represent planning problems.

We use PDDL for symbolic planning.
We define a planning problem $\mathrm{P}$ as a tuple $\langle\sD, s^i, \sS^g\rangle$, where $\sD$ denotes the domain, $s^i$ the initial state, and $\sS^g$ the goal specification, \ie, the set of goal states satisfying the defined goal conditions. A solution to $\mathrm{P}$ is a $T$-step action sequence $a_{1..T}$ such that executing $a_{1..T}$ from $s^i$ leads to a state in $\sS^g$.

The PDDL domain file defines the symbolic action set $\sA$ available in the environment, along with predicates representing object properties. Each action is specified by a string identifier (\eg, \texttt{Putdown}), a list of parameters (\eg, \texttt{?ob}), and its semantics. The semantics $\Phi_a$ of an action $a \in \sA$ include its preconditions (conditions under which the action can be executed) and postconditions (effects describing state changes resulting from the action).

\subsection{Domain Induction}
The domain induction problem has been explored under various setups (Table~\ref{tab:rw_table}).
\citet{arora:hal-02010536} introduced several formulations, including predicting preconditions and postconditions from valid plans or reusable plan fragments.
\citet{Wong2023LearningAP} and~\citet{lasp} focused on completing missing preconditions and effects within a domain file, leveraging existing action semantics from the same or similar domains.
Given a provided problem file, \citet{oswald2024large} demonstrated single-round action semantics generation using LLMs, while \citet{zhu2024languagemodelsinferaction} achieved domain file reconstruction with provided domain predicates in text-based domains by learning from environment feedback.
Unlike LLM+P~\cite{llmp}, which generates PDDL problem files from in-context examples, we infer action semantics directly in domain files in a zero-shot manner.

More recently, InterPreT~\citep{han2024interpret} incorporated natural language feedback from humans during embodied interaction to infer domain predicates and action semantics. Their work addressed complex visual and robotics environments, enabling robots to follow human-specified goals such as \texttt{``stack red block on coaster''}.
In contrast, \ourmethodshort introduces a fully autonomous system for domain induction in visual and robotics settings, without relying on human evaluation or feedback.
\begin{figure}[t]
    \centering
    \includegraphics[width=\linewidth]{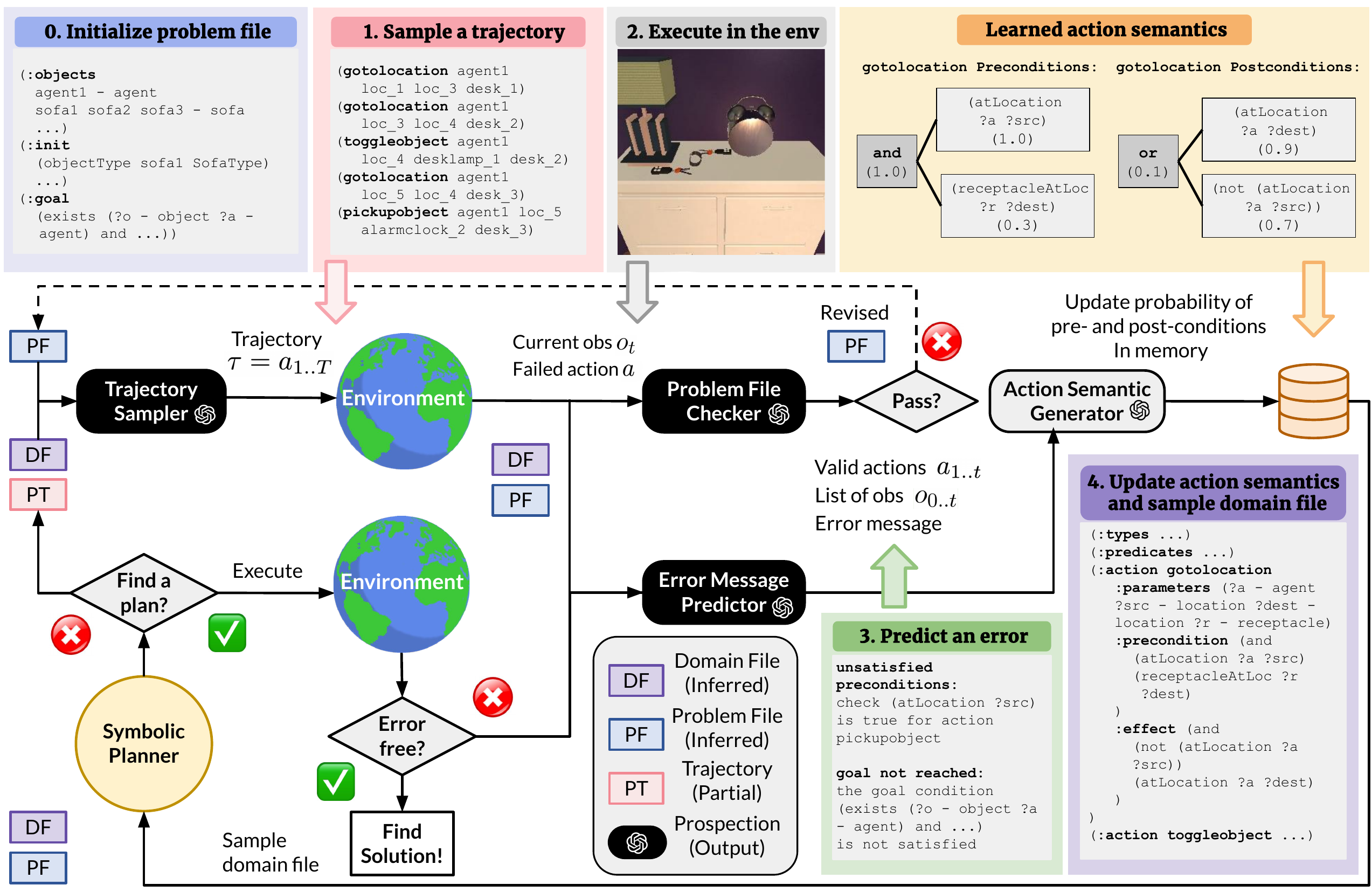}
    \caption{Starting with a problem file initialized from visual input, \ourmethodshort iteratively samples and executes trajectories, predicts error messages, and updates action semantics to refine memory. We maintain a tree-structured belief over possible action semantics for each action, refining these beliefs until a valid plan is found by the symbolic planner that successfully reaches a goal state on execution.}
    \label{fig:vpaslm_method}
\end{figure}

\section{Method}
\label{sec:method}
\ourmethodshort takes as input a natural language task description and a list of domain action signatures---a string name for the action and the type and number of arguments it takes---and predicts both the problem file and the action semantics, \ie, the preconditions and postconditions of actions in the domain file.
The full pipeline of \ourmethodshort is illustrated in Figure~\ref{fig:vpaslm_method}, comprising six main steps.

\textbf{Step 0:} Initialize the problem file using an LLM based on initial, visual observation(s).

\textbf{Step 1:} Given the inferred domain and problem files, sample a trajectory (\ie, a plan) using an LLM. If a partial trajectory is available from the symbolic planner, use it to guide the LLM generation.

\textbf{Step 2:} Execute the sampled trajectory in a simulated or real environment.

\textbf{Step 3:} Correct the problem file based on the execution, or generate execution error message.

\textbf{Step 4:} Predict action semantics from the valid actions and observations and update the memory;

\textbf{Step 5:} Sample a domain file as the current belief and validate it by using the symbolic planner to search for a path from the inferred initial state to the specified goal state; if validation fails, or if the resulting plan fails during execution, return to \textbf{Step 1} with the updated domain file.

For \textbf{Step 1} and \textbf{3}, we perform prospection for $k=5$ steps on LLM output to ensure the first few actions are valid under the inferred domain and problem file. We use GPT-4o as the default LLM unless otherwise specified. All prompts used in this process are detailed in Appendix E.

\begin{figure}[t]
    \centering
    \includegraphics[width=0.9\linewidth]{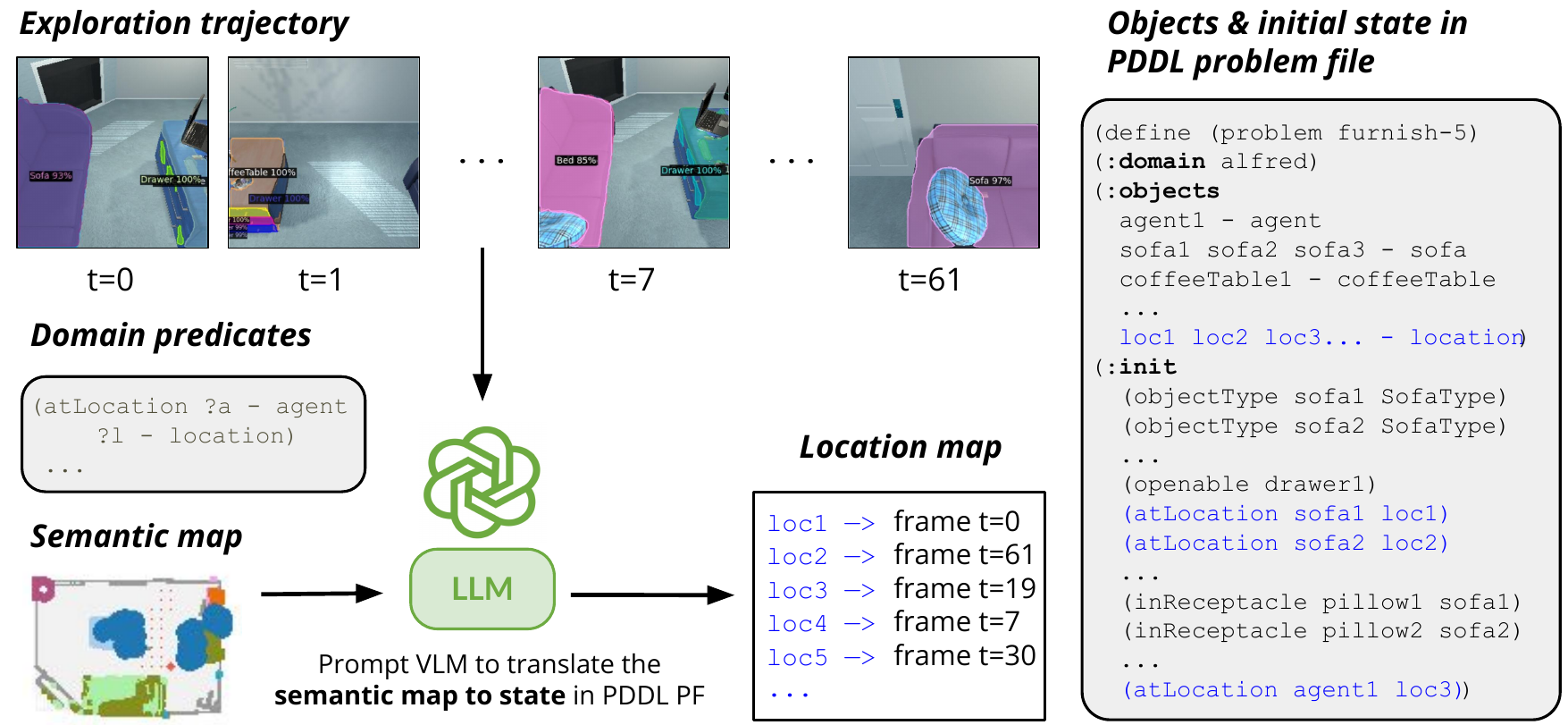}
    \caption{Problem file initialization in the partially observed environment. We first explore the environment to build the semantic map, then use a VLM to translate it to a symbolic initial state.}
    \label{fig:pf_init}
\end{figure}

\paragraph{Problem file initialization.}
A PDDL problem file comprises three core components: objects, initial state, and goal state. We generate these components in two stages due to their differing dependencies. Objects and the initial state depend solely on the environment configuration, while the goal state depends on both the environment and the specified language task.

First, we initialize the objects and initial state based on the environment's observability. 
For fully observed environments, these are generated directly from the initial observation ($o_0$) using a single prompt. For partially observed environments, such as \alfred~\cite{shridhar2020alfred} (illustrated in Figure~\ref{fig:pf_init}), we first perform exploration to build a semantic map using FILM~\citep{min2021film}. 
Subsequently, a Vision-Language Model (VLM) processes this map to generate the PDDL objects, the initial state predicates, and a location map. This location map provides a mapping from abstract location identifiers used in PDDL back to specific coordinates within the interactive visual environment.

Then, we generate the goal state. 
An LLM determines the goal conditions (\eg, one of the goal conditions: \texttt{(exists (?o - object ?a - agent) (and (objectType ?o BookType) (holds ?a ?o)))}) based on the language instruction (\eg, \textit{Take the math book to the night stand}), the previously inferred objects, and the initial state. 
Once generated, this goal state remains fixed. 
To enhance the fidelity of goal generation, especially when a training dataset is available, we employ retrieval-augmented generation (RAG). 
We use BERTScore~\citep{zhang2020BERTScore} to retrieve the top-2 most similar language instructions from the training set as in-context examples during goal generation.

\paragraph{Problem file editing.}
During the execution of a sampled trajectory, we monitor for action failures using environmental feedback, such as observing no change in state or detecting a game restart. 
When a failure occurs, say at time $t$, execution halts. 
The halt signifies that actions $a_{0 \dots t-1}$ were successful, while action $a_t$ failed. 
At this point, an LLM problem file checker analyzes the situation. 
It takes the current observation $o_t$, the attempted action sequence $a_{0 \dots t}$, and the inferred PDDL domain and problem files as input to determine if the failure is due to inaccuracies in the problem file. 
If the checker identifies errors (\eg, missing objects), it proposes corrections. 

Because attempting to correct the domain induction based on failures caused by an erroneous problem file is unreliable, we incorporate the LLM's proposed revisions into the problem file and return to \textbf{Step 1} to re-initiate trajectory sampling with the corrected file.

\paragraph{Action semantics generation.}
Error messages are important for inferring action semantics~\citep{zhu2024languagemodelsinferaction}. We first predict error messages for execution failures, and then use them to generate the action semantics.

An LLM is employed to predict two primary types of errors: 
1) Unsatisfied preconditions, identified when an attempted action fails during execution; and 
2) Unreached goals, identified when a sampled trajectory terminates without achieving the desired goal state. 
To provide informative feedback, particularly when explicit environmental error messages are sparse, the LLM also generates detailed explanations specifying which preconditions were violated or which goal conditions remained unmet.

The list of observations $o_{0 \dots t}$, the sequence of attempted actions $a_{1 \dots t}$, along with the predicted error message and its explanation, serve as input to a separate LLM-based action semantics generator. For cost efficiency, this generator produces the semantics (\eg, pre- and post-conditions) for all domain actions in a single inference call. 
A key distinction from prior work~\citep{zhu2024languagemodelsinferaction}, which restricted logical connectors to only \texttt{and}, is our support for multiple common connectors: \texttt{and}, \texttt{or}, and \texttt{when}. 
The structured output from the LLM is subsequently parsed to extract the formalized action semantics. 

\paragraph{Tree-structured memory update.}
We keep a memory of the hypothesized action semantics. For each action $a$, we store two trees: one for preconditions and one for postconditions. Each tree encodes predicted logical connector nodes and statement nodes.
Each statement node $\phi$ is associated with a belief $p(\phi|a)$, representing the probability that $\phi$ belongs to the semantics of action $a$. This belief is used as a binary sampling probability during each iteration to construct concrete action semantics as input to the symbolic solver.
We define the negation node $\neg\phi$ of a node $\phi$ as a sibling node with the opposite statement, \eg, the negation of ``\texttt{(not (arm-empty))}'' is ``\texttt{(arm-empty)}''. When a node is predicted, the belief of its negation node is reduced accordingly.

When a node $\phi$ is first predicted as part of $a$'s semantics, its belief is initialized to 1 (unless its negation is also predicted). Subsequent updates follow an exponential forgetting rule. Let $\hat{\Phi}_{a, t}$ denote the predicted semantics of $a$ at time step $t$. The update rule for $p_{t+1}(\phi \mid a)$ is:
\begin{align*}
    p_{t+1}(\phi|a) =
    \begin{cases}
    \min(\mathbbm{1}[\phi\in \hat{\Phi}_{a, t+1}] - \alpha\mathbbm{1}[\neg\phi\in \hat{\Phi}_{a, t+1}], 0)  \quad\quad\quad\quad\quad\quad\quad\quad\quad\quad \text{if } p_{t}(\phi|a)=0, \\
    \min(\beta p_{t}(\phi|a) + (1-\beta) \mathbbm{1}[\phi\in \hat{\Phi}_{a, t+1}] - \alpha(1-\beta) \mathbbm{1}[\neg\phi\in \hat{\Phi}_{a, t+1}], 0) \ \ \ \text{otherwise}. \\
    \end{cases}
\end{align*}
Here, $\alpha=0.7$ is the contradiction penalty, and $\beta=0.8$ is the forgetting factor. This update is applied to all nodes $\phi$ predicted in the current step or previously observed, i.e., $\phi \in \bigcup_{t' \in \{1 \dots t+1\}} \hat{\Phi}_{a,t'}$.

\paragraph{Symbolic planner verification.}
We use the \planner~\cite{fast-downward} planner to verify whether a valid plan can be generated from the inferred domain and problem files, with a search time limit of $W=30$ seconds during each \ourmethodshort loop. If \planner fails to produce a complete plan but returns partial trajectories, these are also passed to the trajectory sampler as candidate trajectories.

\begin{table*}[t]
\centering
% \small
\resizebox{\linewidth}{!}{
\tabcolsep 16pt
\begin{tabular}{lrrrrrrr}
    Model & \multicolumn{3}{c}{\alfred} & \multicolumn{2}{c}{\rtfm} & \multicolumn{2}{c}{\overcook} \\
    \cmidrule(lr){2-4}\cmidrule(lr){5-6}\cmidrule(lr){7-8}
    & F1 & SR & GC & F1 & Win & F1 & Win \\
    \midrule
    -- \emph{Plan w/o PDDL} \\
    GPT-4o & - & 31 & 45 & - & 100 & - & 100  \\
    Claude-3.7 & - & 37 & 57 & - & 100 & - & 60 \\
    RoboGPT & - & 60 & 71 & - & - & - & - \\
    \midrule
    -- \emph{Plan after domain induction} \\
    GPT-4o & 71 & 0 & 0 & 85 & 30 & 68 & 0 \\
    Claude-3.7 & 76 & 0 & 0 & 88 & 34 & 75 & 0 \\    
    \midrule
    \ourmethodshort & \bf 91 & \bf 74 & \bf 75 & \bf 100 & \bf 100 & \bf 100 & \bf 100 \\
    \bottomrule        
\end{tabular}
}
\caption{\ourmethodshort achieves strong domain induction (F1) and outperforms other baselines on environment-defined success measures. Direct LLM planning has \emph{no} domain induction F1 score.}
\label{tab:baselines_as}
\end{table*}

\section{Simulation Experiments}
\label{sec:setup}

We evaluate PSALM-V across three simulated visual environments: 
(1) \textbf{\alfred}: An interactive 3D simulation environment where agents must perform complex tasks based on human instructions. We conduct RAG on the training set of \alfred for goal state generation, and evaluate on the unseen validation set.
(2) \textbf{\rtfm}: The \rtfm environment offers a text-based 2D game setting requiring the agent to read textual instructions and act to defeat enemies. To facilitate symbolic planning, we disable the movement of enemies and evaluate on the ``rock\_paper\_scissors'' task.
(3) \textbf{\overcook}: \overcook simulates a multi-agent kitchen scenario where cooperation and real-time decision-making are essential. We evaluate on the ``Forced Coordination'' task.

\begin{figure}[t]
    \centering
    \includegraphics[width=\linewidth]{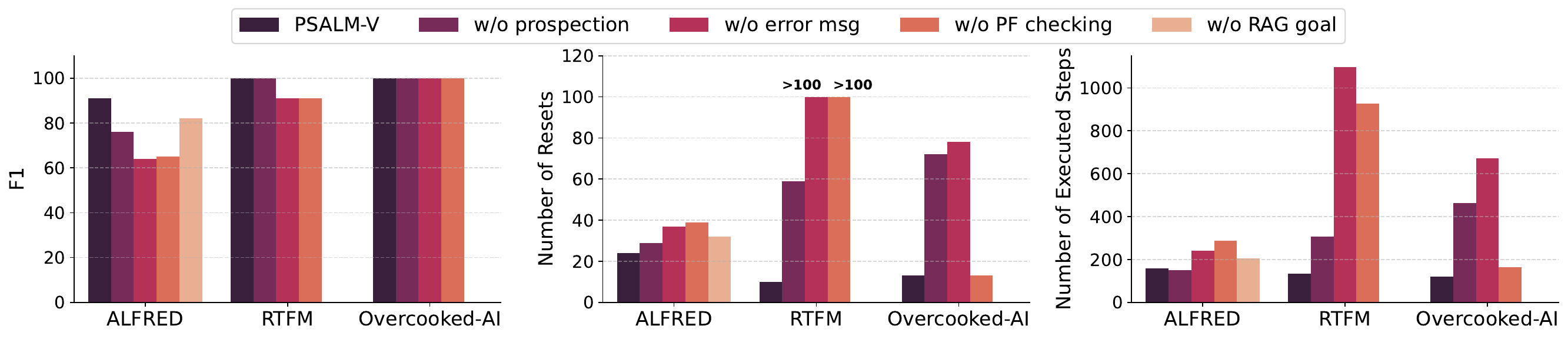}
    \caption{Removing \ourmethodshort\ components such as prospection, error message prediction, and problem file checking increases reset steps and lowers domain induction accuracy (F1).}
    \label{fig:ablation}
\end{figure}

\subsection{Baselines and Evaluation Metrics}
We compare \ourmethodshort to (1) \emph{closed-loop plan generation} and (2) \emph{single-round domain induction} on proprietary models
GPT-4o~\citep{gpt-4o} and Claude-3.7-Sonnet~\citep{claude-3.7-sonnet}, and the current state-of-the-art method RoboGPT~\citep{chen2023robogpt} on \alfred. We also compare with open weight Qwen2.5-72B~\citep{qwen2.5} in Appendix D. 

We evaluate the domain induction learning using three metrics.
(1) \textbf{F1 score}: For each predicted action semantics tree, we list paths from the root to each statement node. \Eg, \texttt{Putdown\_pre-and-(arm-empty)} in the precondition tree of action \texttt{Putdown} from the root to the logical node \texttt{and} down to the leaf \texttt{(arm-empty)}. We compute the F1 score between the predicted and ground truth path lists, with high scores indicating close match.
(2) The number of resets (\textbf{NR}): how many times the environment returns to $s_i$ because planned actions did not reach $\sS^g$.
(3) The number of executed steps (\textbf{NES}): how many total actions were executed during learning.
\subsection{Results and Analysis}
\label{sec:results}

\ourmethodshort successfully recovers the domain and outperforms other baselines on
environment-defined success measures. All components of PSALM-V contribute to accuracy and step efficiency.

% \subsection{Simulated environmentc results}
\paragraph{\ourmethodshort is robust on domain induction, while LLMs fail.}
\ourmethodshort consistently recovers accurate action semantics and problem representations across partially observed and multi-agent environments (Table~\ref{tab:baselines_as}). In contrast, LLMs like GPT-4o and Claude-3.7 never produce valid domain files. 
This finding highlights that one-shot domain inference by LLMs is brittle, whereas \ourmethodshort's iterative refinement and symbolic validation enable reliable planning from incomplete or noisy inputs.

\paragraph{\ourmethodshort enables strong planning after domain induction.}
Table~\ref{tab:baselines_as} shows that LLMs alone struggle with partially observed planning tasks. \ourmethodshort outperforms the best public model, RoboGPT, on \alfred by 14\% in success rate (SR) and 4\% in goal-conditioned completion (GC). Compared to Claude-3.7 and GPT-4o, \ourmethodshort more than doubles the SR on \alfred and achieves consistently robust performance---maintaining a 100\% win rate---on both \rtfm and \overcook.

\paragraph{All components of PSALM-V contribute to accuracy and step efficiency.}
Figure~\ref{fig:ablation} shows that removing components of \ourmethodshort\ leads to reduced F1 scores and increased execution cost, as measured by Number of Resets (NR) and Number of Executed Steps (NES). For example, disabling problem file checking causes NES to spike from 158 to 287 (nearly 2x) in \alfred and from 135 to 928 (nearly 7x) in \rtfm, reflecting inefficient exploration due to undetected problem file errors.

\paragraph{Error messages and prospection are critical for learning from failure.}
Without access to predicted error messages, the model must rely solely on environmental signals, resulting in degraded performance (F1 drops from 91 to 64 in \alfred) and inefficient recovery (NES increases to over 1000 in \rtfm). 
Without prospection, planning becomes less guided, requiring more corrections.

\paragraph{Retrieval-augmented goal inference boosts planning accuracy.}
Ablating RAG-based goal construction reduces both F1 and step efficiency, indicating that retrieved examples help ground goal inference in ambiguous or underspecified tasks.

\begin{figure}[t]
    \centering
    \includegraphics[width=1\linewidth]{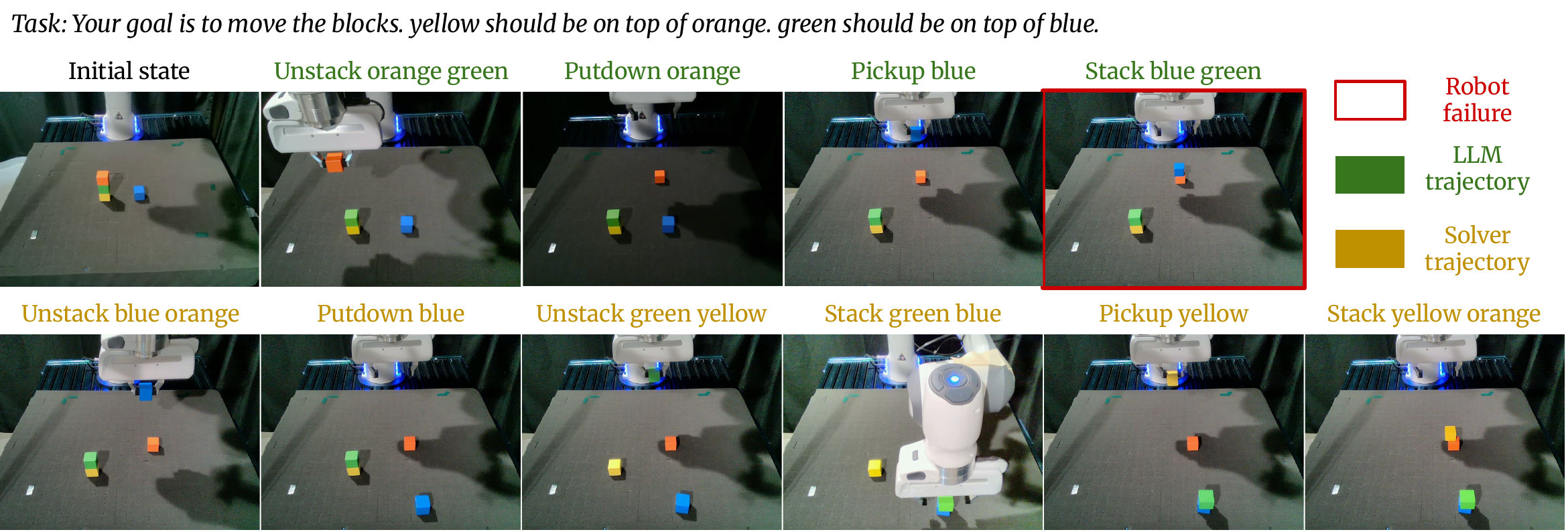}
    \caption{\ourmethodshort remains robust even in the presence of low-level execution failures.}
    \label{fig:psalm-robot}
\end{figure}

\begin{table*}[t]
\centering
% \small
\resizebox{\linewidth}{!}{
\tabcolsep 6pt
\begin{tabular}{lrrrrrrrrrrrr}
    Model & \multicolumn{3}{c}{\bf Task 1: 4 Blocks} & \multicolumn{3}{c}{\bf Task 2: 5 Blocks} & \multicolumn{3}{c}{\bf Task 3: 5 Blocks} & \multicolumn{3}{c}{\bf Overall} \\
    \cmidrule(lr){2-4}\cmidrule(lr){5-7}\cmidrule(lr){8-10}\cmidrule(lr){11-13}
    & F1 & GC & NES & F1 & GC & NES & F1 & GC & NES & F1 & GC & NES \\
    \midrule
    GPT-4o - \textit{Plan} & - & 0 & 4 & - & \bf 50 & 8 & - & 0 & 5 & - & 16.7 & 5.7 \\
    GPT-4o - \textit{Domain} & 91.8 & - & - & 93.9 & - & - & 76.9 & - & - & 87.5 & - & - \\
    \ourmethodshort & 100.0 & \bf 100 & 10 & 100.0 & \bf 50 & 9 & 100 & \bf 50 & 9 & \bf 100.0 & \bf 66.7 & 9.3 \\
    \bottomrule        
\end{tabular}
}
\caption{\ourmethodshort can recover BlocksWorld action semantics completely with an average of 9.3 steps, while GPT-4o is only able to recover it with an average F1 of 87.5\%.}
\label{tab:psalm_robot}
\end{table*}

\section{Real Robot Experiments}
We conduct the 4-action BlocksWorld task~\cite{seipp-et-al-zenodo2022}, where the robot reorganizes a collection of block piles arranged on a table into a specified configuration. The robot has an arm that can hold one block at a time, operating four actions: \texttt{pickup}, \texttt{putdown}, \texttt{stack}, and \texttt{unstack}. We evaluate on 3 BlocksWorld tasks, each with at most 5 colored blocks (See Figure~\ref{fig:psalm-robot} for an example and Appendix A for more). 
Because the task environment is fully-observed and the problem file prediction is accurate, we change to a reset-free setup for the new problem file for the next iteration (\ie, before calling trajectory sampler or symbolic planner) of \ourmethodshort, introducing an additional inference challenge.

We use pretrained skill policies~\cite{ogvla} for each of the above actions (details in Appendix B). The policy predicts two keyframes for each action.
The first keyframe predicts block grasp or release location based on the instruction, and the second keyframe moves the arm to neutral pose above the BlocksWorld environment for the next action execution. The keyframes are achieved via an IK motion planner from Deoxys real-time controller~\cite{zhu2022viola-deoxys} for Franka Emika Panda arm. We use a single front-facing camera.

Table~\ref{tab:psalm_robot} shows that \ourmethodshort successfully recovers the full domain (F1 of 100\%) with an average of 9.3 environment steps and achieves a goal-conditioned completion rate of 66.7\%. In contrast, GPT-4o achieves only partial recovery, with an average F1 score of 87.5\% and a goal-conditioned completion rate of 16.7\%. Notably, \ourmethodshort remains robust even in the presence of low-level execution failures, such as mismanipulating the target object, as illustrated in Figure~\ref{fig:psalm-robot}, where the robot mistakenly stacks the blue cube on the orange cube.
\section{Discussion}
\label{sec:discussion}

We presented \ourmethodshort, a neuro-symbolic framework for planning in partially observed, visually grounded environments without requiring predefined problem files or full observability. By combining LLM-based inference, trajectory execution, and symbolic validation, \ourmethodshort iteratively induces domain semantics and recovers valid PDDL representations.
Our experiments demonstrate strong performance across simulated and real-world settings, showing substantial gains in planning success and semantic induction. These results suggest that integrating language models with symbolic reasoning offers a scalable path toward robust planning in complex environments.
Future work includes scaling to longer-horizon tasks, improving generalization to unseen domains, and incorporating richer feedback modalities beyond language and observation.
\section*{Limitation}
While \ourmethodshort offers a significant step toward automated planning in visual domains, several limitations remain:

\paragraph{Dependency on domain specification.} 
\ourmethodshort requires a set of predefined action names, object types, and predicates to induce action semantics. This partial dependency on prior domain knowledge limits full generalization to novel domains where such abstractions are unavailable. 
Future research could explore using LLMs or pretrained vision-language models to automatically extract candidate predicates and type hierarchies from demonstration data or affordance priors.

\paragraph{Instability on public LLMs.} 
The system's performance depends on accurate error explanation and problem file generation, which can be unstable across less powerful public VLMs (\eg, Qwen-2.5-72B). These models often hallucinate symbolic elements or fail to capture nuanced state transitions, especially under ambiguous feedback.
Future studies could incorporate more systematic verification mechanisms, such as constrained decoding checkers or multi-run tests, to filter or revise outputs from the LLMs, improving reliability across model variants.

\paragraph{Limitations of PDDL formalism.} 
PDDL, while expressive for discrete planning, has known limitations in modeling continuous or dynamic environments, such as those involving time, probability, or numeric reasoning. 
In such settings, symbolic representations may fail to capture fine-grained dynamics (e.g., grasp force, trajectory optimization).
A promising direction is to hybridize PDDL with continuous control modules or extend the symbolic formalism to support numeric fluents and temporal constraints, enabling planning in richer real-world domains.

\bibliography{custom, anthology}

\newpage
\appendix
\section*{Appendix}
%%% Table of contents
\startcontents[appendix]
\addcontentsline{toc}{chapter}{Appendix}
\renewcommand{\thesection}{\Alph{section}} 
\printcontents[appendix]{}{1}{\setcounter{tocdepth}{3}}
\setcounter{section}{0}

\section{Robot BlocksWorld Tasks}
\label{appsec:blocksworld}

\subsection{Task description}

We modified the BlocksWorld IPC tasks~\cite{pddlgendomain} and adapted the tasks to robot domain.
Instead of using symbols \texttt{b1}, \texttt{b2} to represent blocks, we used colors, such as \texttt{green} and \texttt{yellow}, to represent blocks to avoid confusion in object detection.
We selected three representative tasks, with the number of blocks to no more than 5, as we have only 5 different colored blocks.
We describe the three robot tasks, their initial states, goal descriptions, and plans generated by GPT-4o.

\begin{figure}[h]
    \centering
    \includegraphics[width=0.6\linewidth]{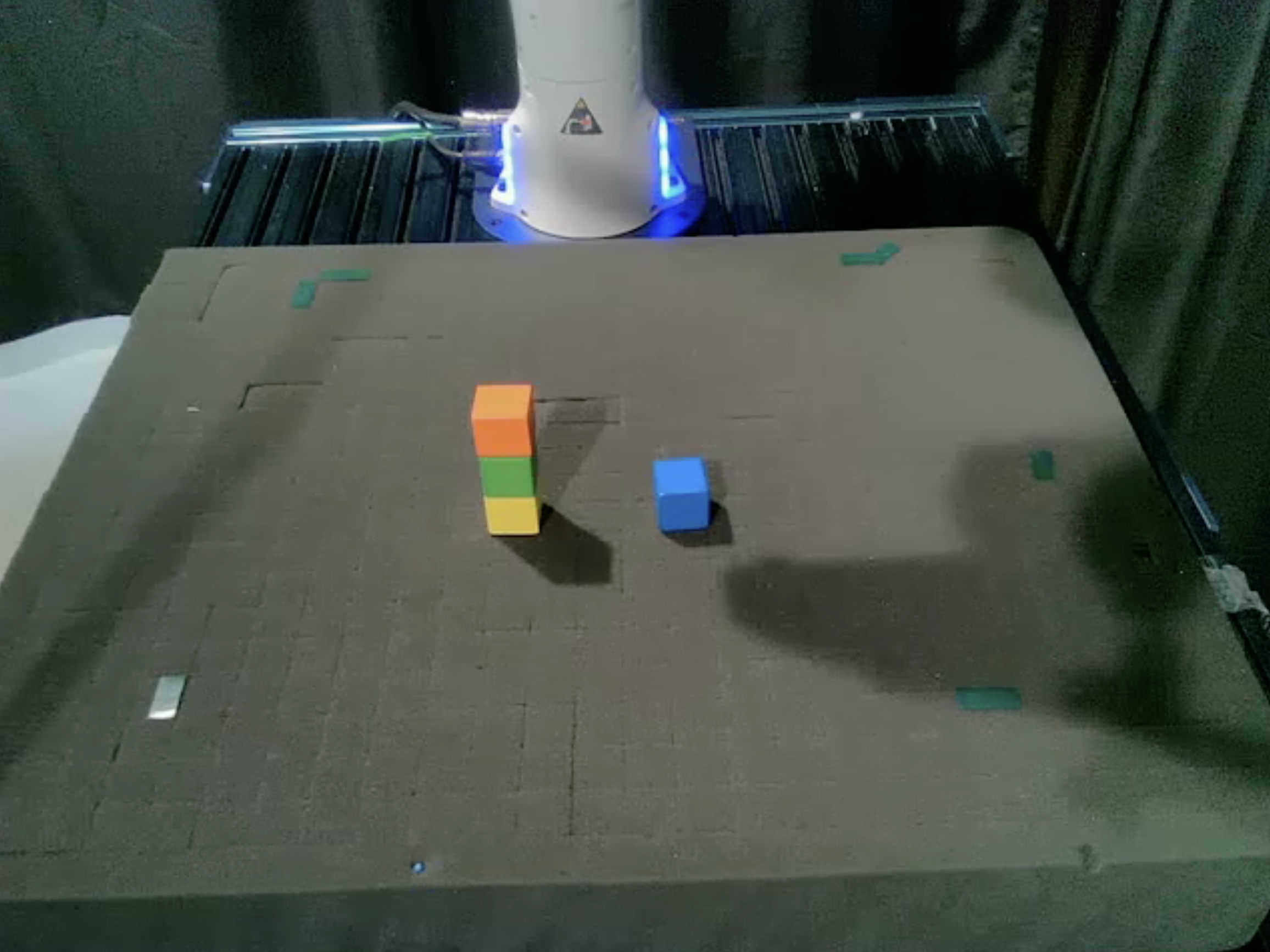}
    \caption{The initial state of robot BlockWorld task 1 on 4 blocks}
    \label{fig:robot_task_1}
\end{figure}

\textbf{Task 1}: Figure~\ref{fig:robot_task_1} show the initial state, the goal description is \textit{Your goal is to move the blocks. yellow should be on top of orange. green should be on top of blue.} The ground truth plan is 

\texttt{(unstack orange green) \\
(putdown orange) \\
(unstack green yellow) \\
(stack green blue) \\
(pickup yellow) \\
(stack yellow orange)}.

And the GPT-4o generated plan is 

\texttt{(unstack orange green) \\
(putdown orange) \\
(unstack green yellow) \\
(stack green blue) \\
(pickup yellow) \\
(stack yellow orange)}.

\begin{figure}[h]
    \centering
    \includegraphics[width=0.6\linewidth]{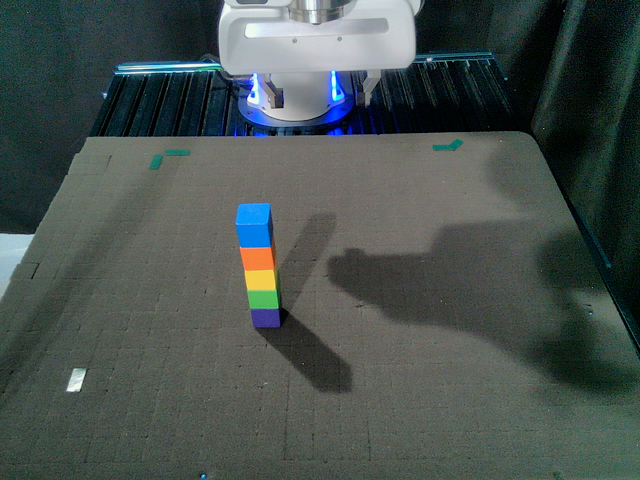}
    \caption{The initial state of robot BlockWorld task 2 on 5 blocks}
    \label{fig:robot_task_2}
\end{figure}

\textbf{Task 2}: Figure~\ref{fig:robot_task_2} show the initial state, the goal description is \textit{Your goal is to move the blocks. orange should be on top of green. green should be on top of purple.} The ground truth plan is 

\texttt{(unstack blue orange) \\
(putdown blue) \\
(unstack orange yellow) \\
(putdown orange) \\
(unstack yellow green) \\
(putdown yellow) \\
(pickup orange) \\
(stack orange green)}.

And the GPT-4o generated plan is 

\texttt{(unstack blue orange) \\
(putdown blue) \\
(unstack orange yellow) \\
(putdown orange) \\
(unstack yellow green) \\
(stack yellow blue) \\
(pickup orange) \\
(stack orange green)}.

\begin{figure}[h]
    \centering
    \includegraphics[width=0.6\linewidth]{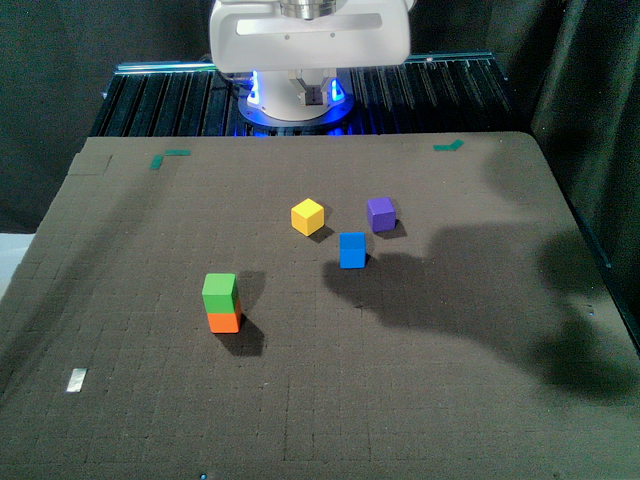}
    \caption{The initial state of robot BlockWorld task 3 on 5 blocks}
    \label{fig:robot_task_3}
\end{figure}

\textbf{Task 3}: Figure~\ref{fig:robot_task_3} show the initial state, the goal description is \textit{Your goal is to move the blocks. green should be on top of purple. 
blue should be on top of green.} The ground truth plan is 

\texttt{(unstack purple yellow) \\
(putdown purple) \\
(unstack yellow green) \\
(putdown yellow) \\
(unstack green orange) \\
(stack green purple) \\
(pickup blue) \\
(stack blue green)}.

And the GPT-4o generated plan is 

\texttt{(unstack purple yellow) \\
(putdown purple) \\
(unstack yellow green) \\
(putdown yellow) \\
(pickup green) \\
(stack green purple) \\
(pickup blue) \\
(stack blue green)
}.

Note that as we limit the number of blocks to 5, this task is not a hard planning task, GPT-4o can sometimes generate correct plans, but they cannot learn action semantics, and cannot recover from low-level failures.

\subsection{Experiment details}

We adapt the \ourmethodshort pipeline to robot tasks by adding a reset-free setup and a step-wise error detector. 

\textbf{Reset-free setup}: Before calling trajectory sampler or symbolic planner, we use GPT-4o to read the current state from image and generate the problem file for the current state. As the modified BlocksWorld task has no more than 5 blocks, the problem file generation by GPT-4o is accurate from manual inspection.

\textbf{Step-wise error detector}: Robot experiment can fail in any action due to low-level manipulation error. Instead of using the LLM error predictor, for each step, we check the action description (\eg, \texttt{(stack green blue))}, and the state transition from the initial state of the action and the end state of the action (\ie, two images) to verify if the state transition matches the action description. If not, we stop the execution and perform action semantic generation.

\section{Robot Learning and Setup Details}
\label{appsec:robot_setup}

Our robot policy~\cite{ogvla} predicts two keyframes for each action.
The first keyframe predicts block grasp or release location based on the instruction, and the second keyframe moves the arm to neutral pose above the BlocksWorld environment for the next action execution. The keyframes are achieved via an IK motion planner from Deoxys real-time controller~\cite{zhu2022viola-deoxys} for Franka Emika Panda arm. We use a single front-facing camera. We omit the details on how the policy works.

\section{Simulation Environment Details}
\label{appsec:sim_env}

\subsection{Simulation Environments}

\begin{figure}[t]
    \centering
    \includegraphics[width=\linewidth]{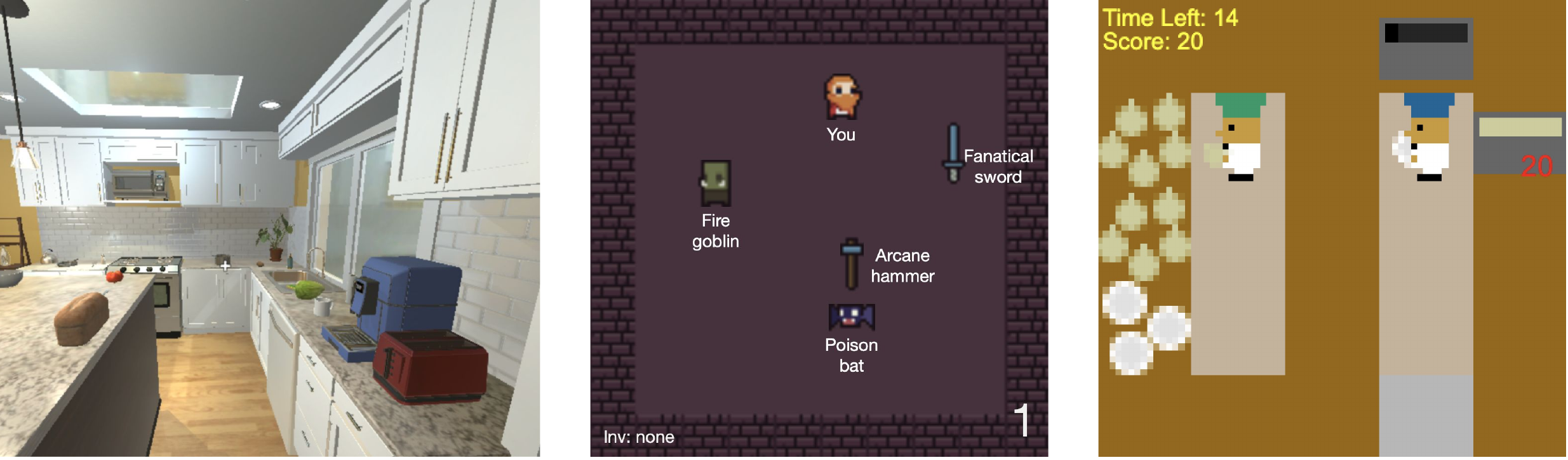}
    \caption{Simulated envs: (left) \alfred~\cite{shridhar2020alfred}, (middle) \rtfm~\cite{zhong2021rtfmgeneralisingnovelenvironment}, (right) \overcook~\cite{carroll2019utility}.}
    \label{fig:psalmv-simenv}
\end{figure}

In Figure~\ref{fig:psalmv-simenv}, we show the visualization of three environments.

\paragraph{\alfred.}
The \alfred benchmark is a large-scale dataset for evaluating embodied agents on instruction-following tasks in interactive environments. It consists of expert demonstrations paired with natural language directives, divided into distinct training, validation, and test sets. The dataset is carefully split to test generalization, with both seen and unseen splits in the validation and test sets.

The training set contains 21,023 annotated trajectories spanning 108 distinct scenes. The validation set includes 820 annotations over 88 scenes for the seen split and 821 annotations over 4 scenes for the unseen split. The test set similarly includes 1,533 seen examples from 107 scenes and 1,529 unseen examples from 8 scenes. 

\alfred has a total of 10 actions in the domain file (we remove the unnecessary actions in the domain file, such as \texttt{look} and \texttt{help}): \texttt{SliceObject}, \texttt{ToggleObject}, \texttt{CoolObject}, \texttt{HeatObject}, \texttt{CleanObject}, \texttt{PutObject}, \texttt{PickupObject}, \texttt{CloseObject}, \texttt{OpenObject} and \texttt{GotoLocation}.
As the testing data has no ground truth domain file to compare the F1 score, for a fair comparison with the previous works we use the average score on the validation unseen data for evaluation. The evaluation measure in \alfred is the success rate (SR) and the goal-conditioned completion (GC).

\paragraph{\rtfm.}
The \rtfm benchmark is designed to test an agent's ability to jointly reason over a structured goal, a symbolic document describing environment-specific rules, and first-person environment observations. Each episode provides a unique setting with a natural language document that defines relationships among monsters, item types, and their interactions. The agent must parse this document, identify relevant entities in the environment, and select appropriate actions to achieve the goal.

We use the ``rock\_paper\_scissors'' task, which has 4 actions \texttt{up}, \texttt{down}, \texttt{left} and \texttt{right}. The original \texttt{stop} action is not meaningful as we force the environment to be static for our experiment. The evaluation measure in \rtfm is the win rate (Win), which is originally the rate of winning rate over multiple trails. We incorporate the efficiency measure in it and redefine the win rate as the winning score over optimal winning score in 100 steps.

\paragraph{\overcook.}: \overcook is a multi-agent planning and collaboration environment designed to evaluate cooperative human-AI task performance. 
Inspired by the popular video game \textit{Overcooked}, the benchmark involves two agents working together in a shared kitchen environment to prepare and deliver soups. Each soup requires placing up to three ingredients into a pot, allowing it to cook, and then serving it. Agents must coordinate actions such as ingredient collection, pot monitoring, and delivery to maximize reward within a limited time frame.

To enforce multi-agent evaluation of \ourmethodshort, we evaluation on the ``Force Coordination'' layout, with 6 actions \texttt{up}, \texttt{down}, \texttt{left}, \texttt{right}, \texttt{pick-up} and \texttt{put-down}. \overcook is evluated on the win rate, similar as in \rtfm.

\subsection{Details on prospection}
Even though the LLM is prompted with the current beliefs about action preconditions, it can still produce action sequences that violate those rules. To address this, we introduce trajectory prospection --- a mechanism that allows the system to simulate ahead using its symbolic understanding of the world, without yet interacting with the actual environment.

Given a proposed trajectory $\mathbf{\tau} = a_{1:T}$, we perform a symbolic check on the first $k$ steps to ensure they are consistent with the current action semantics. If any action among $a_{1:k}$ violates its preconditions according to the current belief, we reject that action and repeatedly sample a new one until we find a valid alternative. This process continues until we obtain a prefix of $k$ valid steps. If all of $a_{1:k}$ are valid from the start, we keep the original trajectory $a_{1:T}$ as-is. In practice, we set $k=5$ to ensure that the trajectory begins with a reliable sequence of five valid actions before execution.

\section{Additional Experiments}
\label{appsec:add_exps}

\subsection{Definition of evaluation measures}
Let $\hat{\Phi}_a$ denotes the predicted action semantics of action $a$ and ${\Phi}_a$ denotes the ground truth action semantics of action $a$. The F1 used in our experiment is defined as
\begin{align*}
& precision=\sum_{a\in\sA}|\hat{\Phi}_a\cap\Phi_a|/\sum_{a\in\sA}|\Phi_a| \\
& recall=\sum_{a\in\sA}|\hat{\Phi}_a\cap\Phi_a|/\sum_{a\in\sA}|\hat{\Phi}_a| \\
& \text{F1}=2 \cdot precision \cdot recall / (precision + recall)
\end{align*}

\subsection{Exact number of \ourmethodshort ablations}
\begin{table*}[t]
\centering
% \small
\tabcolsep 11pt
\resizebox{\linewidth}{!}{
\begin{tabular}{lrrrrrrrrr}
    Model & \multicolumn{3}{c}{\alfred} & \multicolumn{3}{c}{\rtfm} & \multicolumn{3}{c}{\overcook} \\
    \cmidrule(lr){2-4}\cmidrule(lr){5-7}\cmidrule(lr){8-10}
    & F1 & NR & NES & F1 & NR & NES & F1 & NR & NES \\
    \midrule
    \ourmethodshort & 91 & 24 & 158 & 100 & 10 & 135 & 100 & 13 & 119 \\
    \midrule
    \emph{w/o prospection} & 76 & 29 & 151 & 100 & 59 & 306 & 100 & 72 & 463 \\
    \emph{w/o error message} & 64 & 37 & 241 & 91 & $>$100 & 1097 & 100 & 78 & 672 \\
    \emph{w/o PF checking} & 65 & 39 & 287 & 91 & $>$100 & 928 & 100 & 13 & 165 \\
    \emph{w/o RAG goal} & 82 & 32 & 205 & - & - & - & - & - & - \\
    \bottomrule        
\end{tabular}
}
\caption{Removing components such as prospection, error message prediction, or problem file checking increases revision steps and lowers F1. RAG goal is only applied on \alfred as it has a training set.}
\label{tab:ablation}
\end{table*}

We show the exact number of \ourmethodshort ablations in Table~\ref{tab:ablation} (\cf Figure~\ref{fig:ablation}).

\paragraph{Each PSALM-V component enhances accuracy and step efficiency.}
Table~\ref{tab:ablation} illustrates that removing any component from \ourmethodshort\ results in lower F1 scores and higher execution costs, as indicated by the Number of Resets (NR) and Number of Executed Steps (NES). For instance, disabling problem file checking causes NES to surge from 158 to 287 in \alfred\ and from 135 to 928 in \rtfm, indicating inefficient exploration due to unresolved problem file issues.

\paragraph{Prospection and error feedback are essential for effective failure recovery.}
Eliminating predicted error messages forces the model to depend solely on environment feedback, leading to performance drops (F1 falls from 91 to 64 in \alfred) and inefficient retries (NES exceeds 1000 in \rtfm). Similarly, without prospection, the model lacks foresight in planning, resulting in more frequent corrections.

\paragraph{RAG-based goal inference improves both planning precision and efficiency.}
Removing retrieval-augmented goal construction leads to declines in both F1 and step efficiency, highlighting the value of retrieved examples in guiding goal inference, especially under ambiguity or limited task specification.

\subsection{Extension to public models}
\begin{table*}[t]
\centering
% \small
\resizebox{\linewidth}{!}{
\tabcolsep 11pt
\begin{tabular}{lrrrrrrrrr}
    Public Module & \multicolumn{3}{c}{\alfred} & \multicolumn{3}{c}{\rtfm} & \multicolumn{3}{c}{\overcook} \\
    \cmidrule(lr){2-4}\cmidrule(lr){5-7}\cmidrule(lr){8-10}
    Qwen-72B & F1 & NR & NES & F1 & NR & NES & F1 & NR & NES \\
    \midrule
    None & 83 & 26 & 179 & 100 & 10 & 135 & 100 & 13 & 119 \\
    \midrule
    TS & 76 & 37 & 245 & 100 & 21 & 279 & 100 & 26 & 283 \\
    ASG & 70 & 42 & 306 & 100 & 52 & 513 & 100 & 32 & 341 \\
    PFC & 34 & 89 & 688 & 91 & >100 & 832 & 100 & 55 & 589 \\
    EMP & 61 & 41 & 299 & 100 & 35 & 320 & 100 & 29 & 303 \\
    \midrule
    TS \& ASG & 70 & 55 & 371 & 100 & 53 & 524 & 100 & 31 & 330 \\
    TS \& ASG \& EMP & 59 & 87 & 576 & 87 & >100 & 892 & 95 & >100 & 723 \\
    \bottomrule        
\end{tabular}
}
\caption{Replacing GPT-4o modules in \ourmethodshort with a public model Qwen-2.5-72B: results show performance degradation in F1 and step efficiency (NR, NES) under partial replacements, especially the problem file checker. TS: trajectory sampler; ASG: action semantics generator; PFC: problem file checker; EMP: error message predictor.}
\label{tab:public}
\end{table*}

From initial evaluation, we noticed that replacing GPT-4o with a public vision-language model for the whole \ourmethodshort system will break the system and result in a poor (<30) F1 score, even with careful parsing strategy to avoid syntactical errors.
To systematically study the generalizability to public models, we replace the GPT-4o LLM modules in \ourmethodshort by a powerful public vision-language model, Qwen-2.5-72B~\cite{qwen2.5}. 
We list the results in Table~\ref{tab:public}. Note that for time efficiency we evaluate on 10\% of the validation unseen data in \alfred. The results show performance degradation in F1 and step efficiency (NR, NES) under partial replacements, especially
the problem file checker. We demonstrate that public LLMs still face limitations in structured planning tasks.

\paragraph{Trajectory sampling with Qwen-72B preserves planning accuracy, but increases execution cost.}
When replacing only the trajectory sampler (TS) with Qwen-72B, the F1 score remains high across all environments (100 in \rtfm and \overcook, and 76 in \alfred). However, NES increases significantly, for example, from 179 to 245 in \alfred and from 135 to 279 in \rtfm, indicating that the sampled plans are less step-efficient, possibly due to reduced commonsense priors in trajectory generation. 
These results suggest that while Qwen-72B can propose reasonable action sequences, it lacks GPT-4o's ability to optimize for minimal correction and retry cycles.

\paragraph{Replacing the action semantics generator leads to semantic misalignment.}
Substituting the action semantics generator (ASG) with Qwen-72B causes performance drops in both accuracy and step efficiency. The F1 score decreases to 70 in \alfred and 100 to 92 in \overcook. NES also increases substantially (\eg, 306 in \alfred and 513 in \rtfm), indicating that the inferred semantics become noisier or less aligned with the environment's execution logic. This results in higher failure rates and more corrective actions.

\paragraph{Problem file checking (PFC) is essential for reliable exploration.}
The most significant degradation occurs when replacing the problem file checker (PFC). In \alfred, F1 drops to 34 and NES surges to 688, while in \rtfm and \overcook, NES exceeds 800 and 580 respectively. 
More importantly, Qwen-72B frequently makes unnecessary edits to the problem file when the problem file is correct, which induces additional complexity to the domain induction process.
The phenomenon occurs less in less complex and smaller environment, such as \overcook (5x5 grid).
This results demonstrate the critical role of robust problem validation in reducing resets and guiding the planner away from invalid trajectories.

\paragraph{Error message prediction supports efficient learning from failures.}
Replacing the error message predictor (EMP) also negatively affects performance. While the F1 scores remain relatively moderate (e.g., 61 in \alfred and 100 in \rtfm), the number of executed steps increases significantly—NES rises to 299 in \alfred and over 300 in \rtfm. This implies that without accurate failure feedback, the system must rely solely on environment-level signals, which are often sparse or delayed, resulting in inefficient recovery and increased exploratory retries.

\paragraph{Combined replacement of multiple modules compounds errors.}
When replacing multiple GPT-4o modules with Qwen-72B simultaneously, the degradation becomes more severe. For instance, replacing TS \& ASG results in an F1 of 70 in \alfred but increases NES to 371; further adding EMP causes F1 to drop to 59 and NES to spike to 576. In \rtfm and \overcook, NES consistently exceeds 700 with F1 dropping below 90. These compounding effects confirm that while Qwen-72B can fulfill individual roles moderately well, its performance suffers dramatically when expected to coordinate multiple interdependent reasoning tasks.

\section{Prompt Templates}
\label{appsec:prompt}

\begin{figure}[t]
    \centering
    \includegraphics[width=0.8\textwidth]{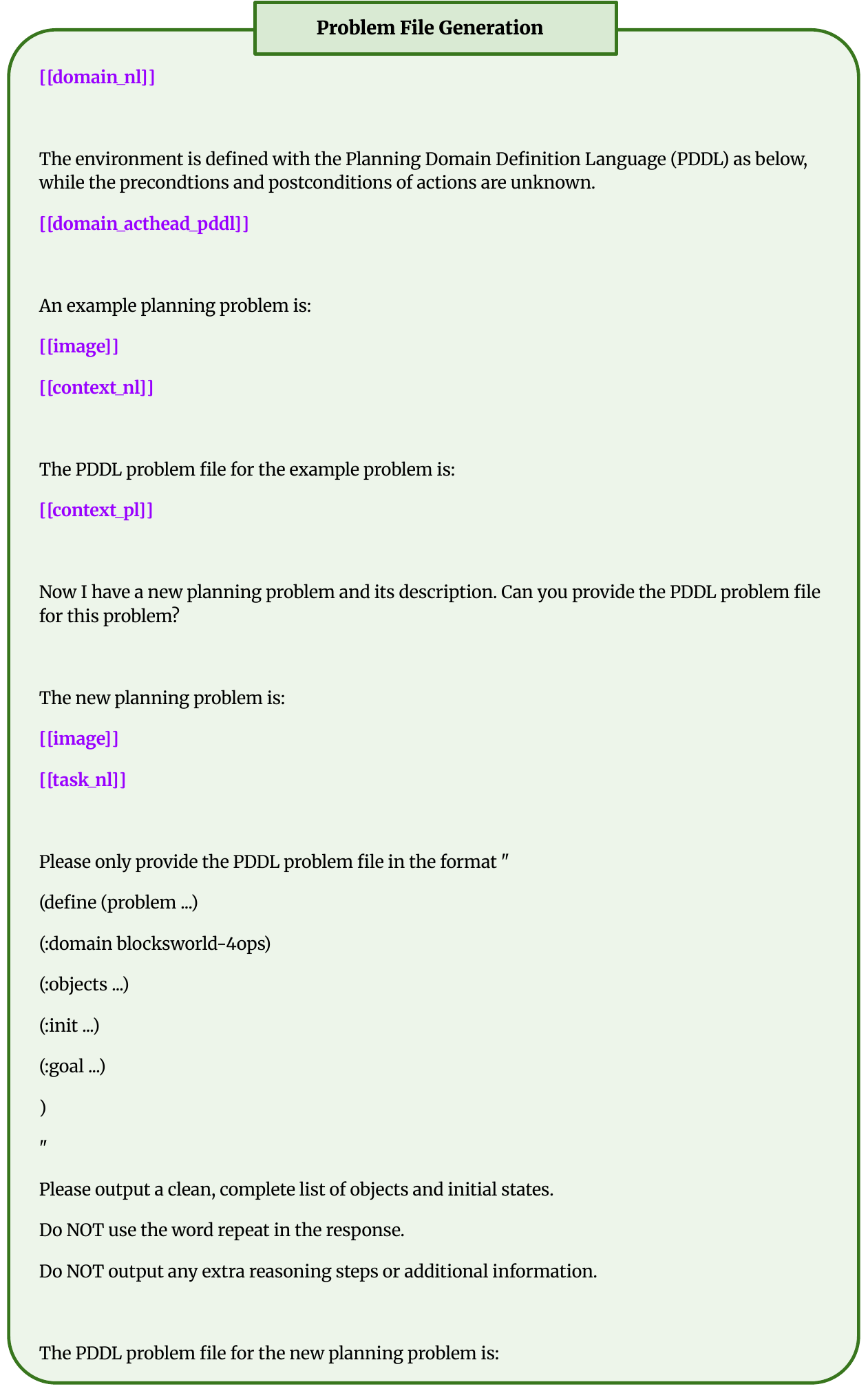}
    \caption{Prompt for PDDL problem file generation.}
    \label{fig:problem_prompt}
\end{figure}

\begin{figure}[t]
    \centering
    \includegraphics[width=0.8\textwidth]{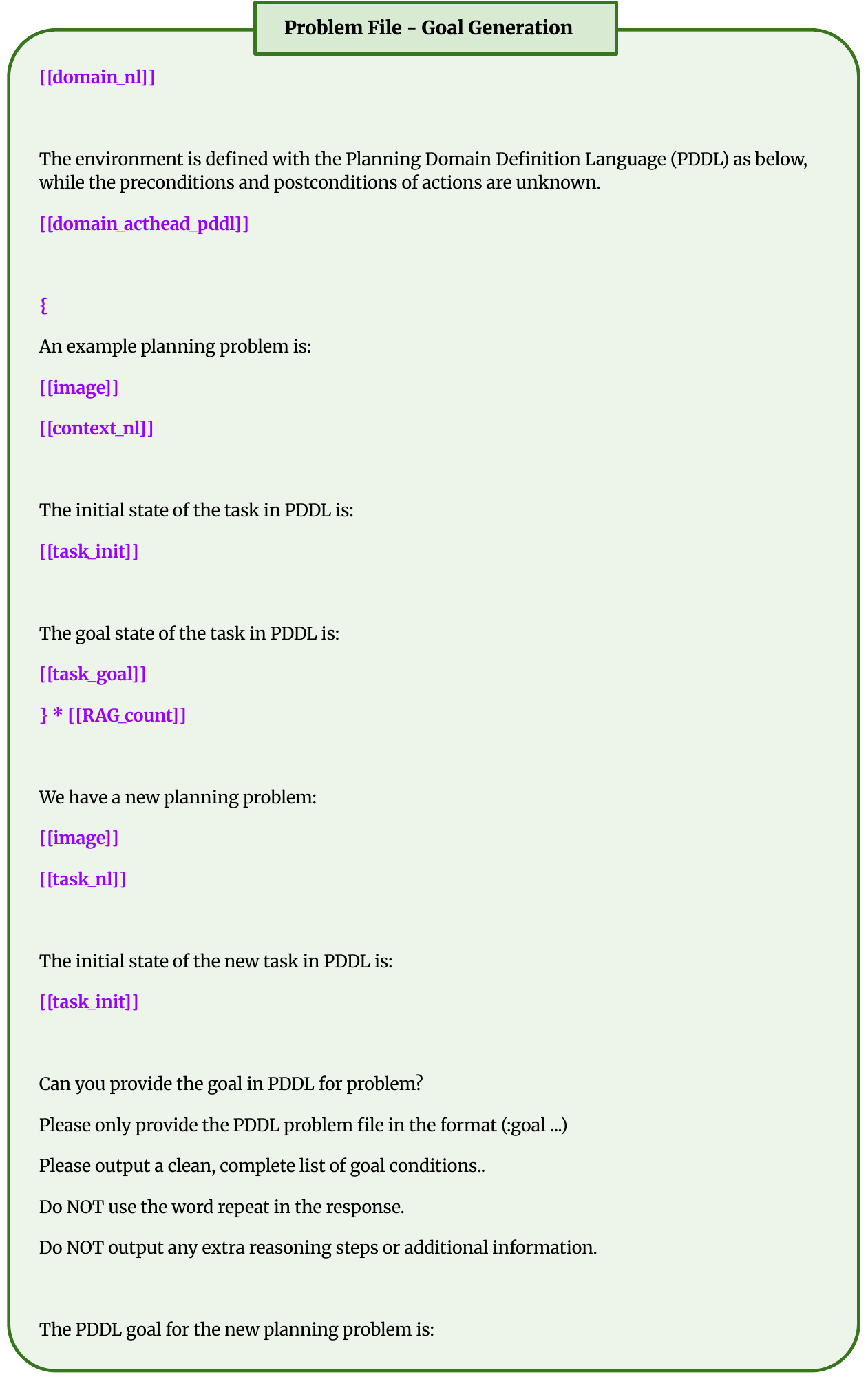}
    \caption{Prompt for PDDL goal generation.}
    \label{fig:problem_goal_prompt}
\end{figure}

\begin{figure}[t]
    \centering
    \includegraphics[width=0.8\textwidth]{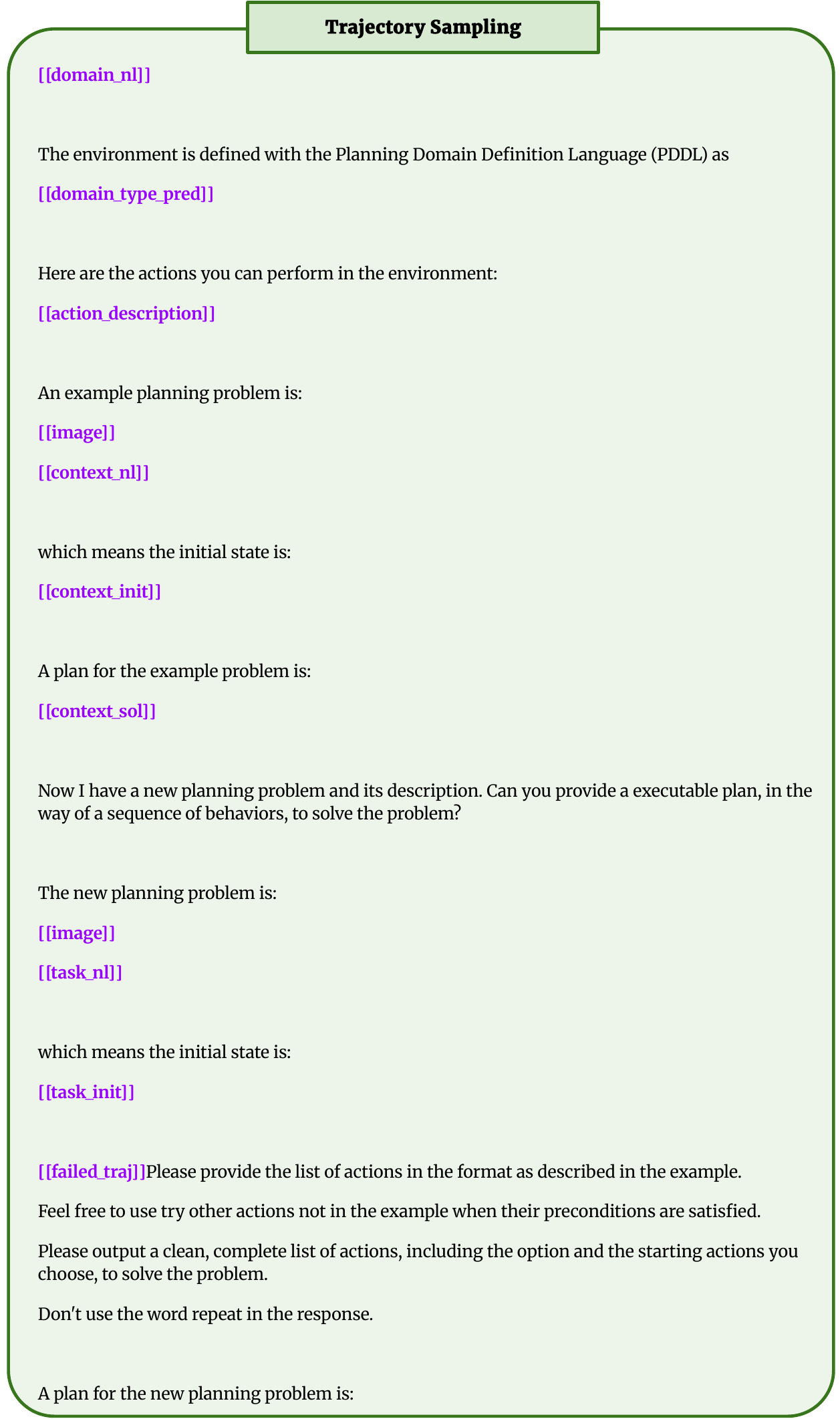}
    \caption{Prompt for trajectory sampling.}
    \label{fig:trajsam_prompt}
\end{figure}

\begin{figure}[t]
    \centering
    \includegraphics[width=0.8\textwidth]{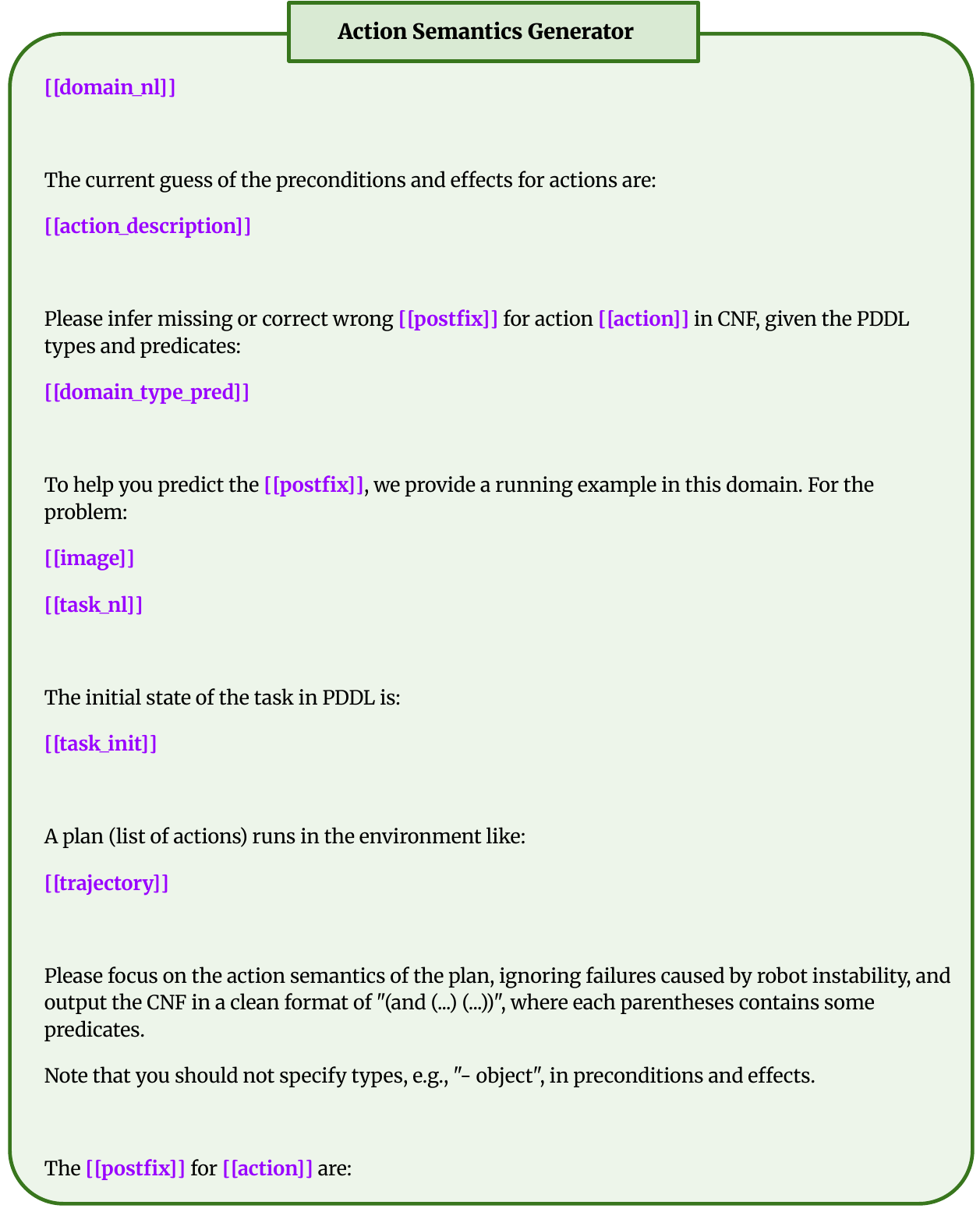}
    \caption{Prompt for action semantics generator.}
    \label{fig:asg_prompt}
\end{figure}

\begin{figure}[t]
    \centering
    \includegraphics[width=0.8\textwidth]{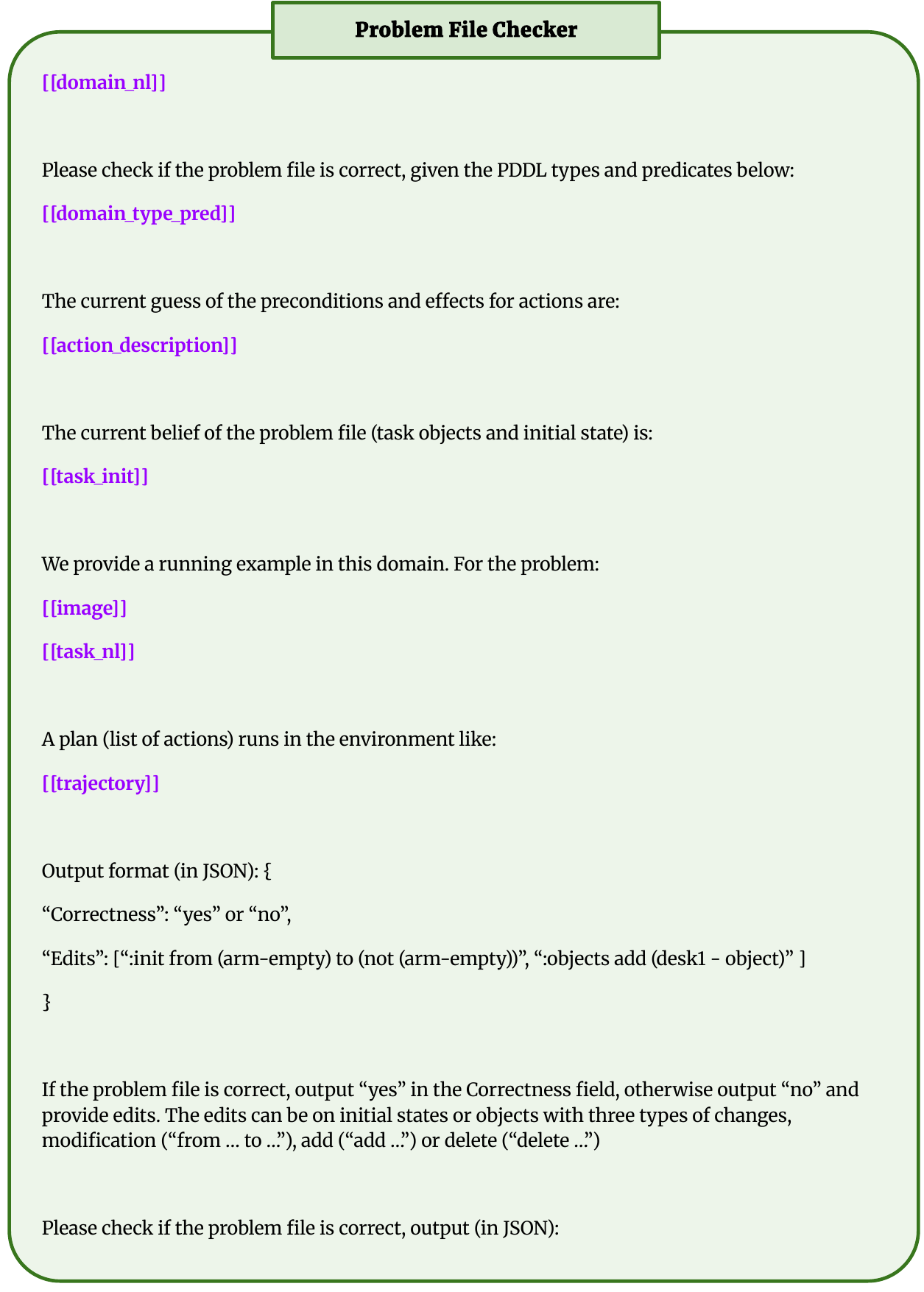}
    \caption{Prompt for problem file checker.}
    \label{fig:pfc_prompt}
\end{figure}

\begin{figure}[t]
    \centering
    \includegraphics[width=0.8\textwidth]{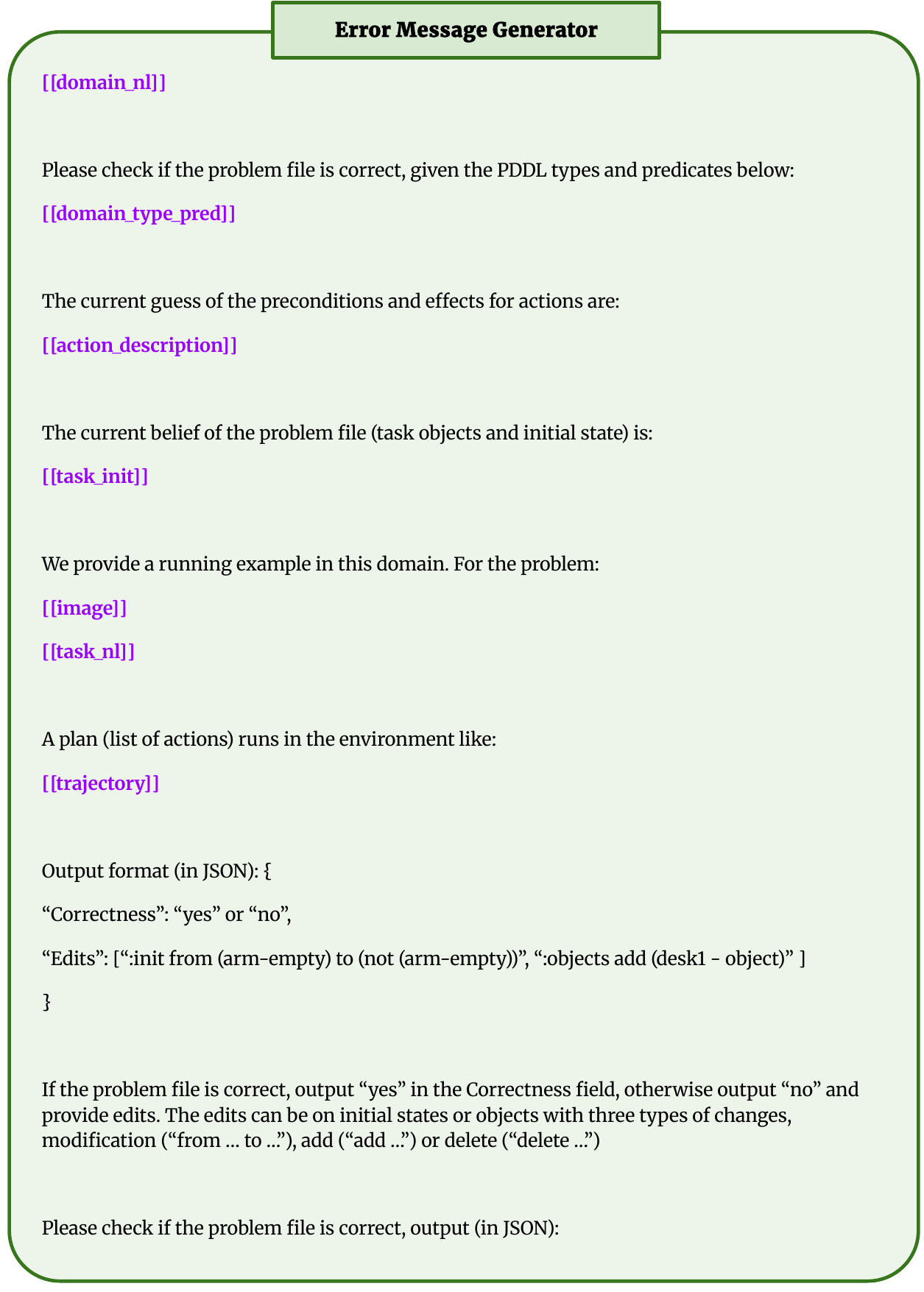}
    \caption{Prompt for error message predictor.}
    \label{fig:emp_prompt}
\end{figure}

Here, we list the prompt templates for PDDL problem file generation (Figure~\ref{fig:problem_prompt}), PDDL goal generation (Figure~\ref{fig:problem_goal_prompt}) trajectory sampling (Figure~\ref{fig:trajsam_prompt}), action semantics generator (Figure~\ref{fig:asg_prompt}), 
problem file checker (Figure~\ref{fig:pfc_prompt}) and error message predictor (Figure~\ref{fig:emp_prompt}).

\end{document}